\documentclass{article}
\usepackage{ijcai19}

\usepackage{times}
\usepackage{soul}
\usepackage{url}
\usepackage[hidelinks]{hyperref}
\usepackage[utf8]{inputenc}
\usepackage[small]{caption}
\usepackage{graphicx}
\usepackage{amsmath}
\usepackage{booktabs}
\usepackage{times}  
\usepackage{helvet}  
\usepackage{courier}  
\usepackage{url}  
\usepackage{graphicx}  
\usepackage{amsmath, amssymb, amsfonts,semantic,colortbl,mathrsfs,stmaryrd}
\usepackage{multirow}
\usepackage{subfigure}
\usepackage[ruled,lined,boxed,linesnumbered,noend]{algorithm2e}

\newcommand{\network}[1]{\mathfrak{#1}}

\newcommand{\ch}[1]{\mathrm{CH}{(#1)}}
\newcommand{\ach}{\mathrm{CH}}

\title{On  Geometric Structure of  Activation Spaces in Neural Networks}

\author{
 Yuting Jia$^1$\and
 Haiwen Wang$^1$\and
 Shuo Shao$^{1,*}$ \and
 Huan Long$^{1,*}$\and
 Yunsong Zhou$^1$\and
 Xinbing Wang$^1$\and
 \affiliations
 $^1$Shanghai Jiao Tong University 
\emails
\{hnxxjyt, wanghaiwen, shuoshao, longhuan, zhouyunsong, xwang8\}@sjtu.edu.cn
}

\begin{document}

\maketitle

\begin{abstract}
In this paper, we investigate the geometric structure of activation spaces of fully connected layers in neural networks and then show applications of this study.
We propose an efficient approximation algorithm to characterize the convex hull of massive points in high dimensional space. 
Based on this new algorithm, four common geometric properties shared by the activation spaces are concluded, which gives a rather clear description of the activation spaces.
We then propose an alternative classification method grounding on the geometric structure description, which works better than neural networks alone.
Surprisingly, this data classification method can be an indicator of overfitting in neural networks.
We believe our work reveals several critical intrinsic properties of modern neural networks and further gives a new metric for evaluating them.

\end{abstract}

\section{Introduction}\label{sec-intro}

In recent years, neural networks have been widely used in areas including pattern recognition and data classification \cite{schmidhuber2015deep}.
While they have delivered the state-of-the-art accuracy on many artificial intelligence tasks, little of their intrinsic mechanism has been revealed to people.
Moreover, if we only use accuracy to evaluate neural networks, some fundamental information hidden behind the complicated architectures may be neglected.
By taking the output of each layer in the neural network as a vector, the set of vectors for each layer will form a space which we call activation space.
Correspondingly, the vectors in the space are called activation vectors.
Activation spaces contain the intrinsic mathematical information learnt by the corresponding neural network.
To gain better understanding of neural networks, we try to uncover the geometric structure of their activation spaces.

Although one straightforward attempt to understand the structure of activation spaces would be dimension reduction or even data visualization, and there have been a lot of related works, such as t-SNE \cite{Maaten2014}.
However, dimension reduction methods may lead to complications (like overlapping) in data points, which do not happen in the original high dimensional space.
In \cite{raghu2017svcca}, the researchers study the space formed by all input data and give a method to measure the intrinsic dimensionality of layers.
In \cite{yu2018curvature}, the authors study the local dimension of data manifolds which could be seen as trying to reveal the microstructure of activation spaces in each layer.
The most recent work is \cite{wang2018towards}, the authors take activation spaces as the basis to do the comparison between two neural networks and focus on the matching relation among neurons. 
Comparatively, our work intends to uncover the fundamental geometric structure of the original activation space.
For this goal, we study the convex hull of activation vectors of each class in every fully connected layer.

In this paper, we focus on the neural networks for classification.
We want to figure out how the geometry of all data of the same class varies from one layer to the next in activation spaces, and what is the geometric difference among data of different classes in each activation space.
Now we only study the neural networks with fully connected layer and Softmax layer, for we believe the fully connected structure is the most representative and fundamental factors in neural networks considering activation spaces.
Nevertheless, it is clear that our method and ideas can be extended to more general situations (including convolutional layers, or neural networks of different architectures) as long as the outputs can be represented by vectors.

On the one hand, we capture the high-level structure of the activation spaces by characterizing the its convex hull.
As the exact convex hull of high dimensional data is computationally infeasible to get,
we propose a new approximation algorithm which works well for the data in activation spaces of neural networks.
In addition, to figure out how the datasets align in each activation spaces, we explore the activation spaces by studying some basic geometric properties, such as the distance distributions of data of the same and different classes. Some interesting conclusions come to light in this part.

On the other hand, we show the above approximate convex hull algorithm can actually do more than it appears.
Combining it with our knowledge of the activation spaces, we derive a new classification method and show its power.
Besides accuracy improvement, this new method is instructive as it can also serve as an indicator of overfitting in neural networks.
Moreover, it turns out that some of the results in this part match some previous hypothesis and theories.

In summary, the contributions of this paper are threefold:
\begin{itemize}
  \item We try to characterize the geometric structure of activation spaces in neural networks by the convex hull and give four interesting results, along with a new efficient approximate convex hull algorithm.
  \item Based on the understanding of the geometric structure of activation spaces, we propose the nearest convex hull classification method, which surprisingly outperforms the more dedicated neural network in several cases.
  \item We give a new reasonable metric for deciding the overfitting, and a possible way to prune neural networks.
\end{itemize}

The remainder of the paper is organized as follows. \S\ref{sec-actSpace} gives the mathematical definition of activation vectors in neural networks.
To get the convex hull of activation vectors of different classes, we build a new approximate algorithm which works efficiently for the high dimensional vectors.
\S\ref{sec-STRU} contains our main results about the geometric structure of the activation spaces.
We show that though convex hull may seem to be an oversimple estimation of the distribution of activation vectors, further experiments show a surprising contrary.
Then based on this understanding of the geometric structure, in \S\ref{sec-CCH} we implement the closest convex hull algorithm to every activation space and show an instructive difference considering the original neural networks.
\S\ref{sec-related} addresses some related work.
Finally, \S\ref{sec-conclusion} concludes the whole paper with more discussion and future work.

\section{Activation Space and Its Approximation} \label{sec-actSpace}

In this paper, we take the convention that all vectors are column vectors and denoted by lower-case bold letters like $\mathbf{x}$.
Calligraphic letters are used to represent sets like $\mathcal{X}$.
For simplicity, the corresponding normal letters are used to denote their size like $X=|\mathcal{X}|$.

\subsection{Activation spaces in neural networks}\label{subsec-activationSpace}
One of the tempting explanation for the success of deep neural networks is that they are highly non-linear and nonconvex \cite{nelles2013nonlinear}.
Here instead of taking this well-accepted hypothesis for granted, we carry out our work in a different way: we consider the convex hull of vectors in activation spaces and then try to characterize their geometric structure.

The convex hull is one of the most important conceptions in computational geometric.
It is a critical notion widely used in data description and analysis.
For a given set of points $\mathcal{S}$, its convex hull is defined as the smallest convex set that contains $\mathcal{S}$.
Convex hull captures the high-level information and also some intrinsic properties of the datasets.
When the dimension is low, the traditional algorithms like Graham’s scan method \cite{Graham1972} and the gift wrapping algorithm \cite{Donald1970,Jarvis1973} will work in $O(S\log S)$ time.
However as it comes to high dimensional data, the exact representation of the convex hull becomes computationally infeasible to get with a lower bound relying exponentially on the dimension $d$ \cite{DwyerPhD1988}.
As dimensions in neural networks are usually very high, it is natural to pursue an efficient approximation algorithm instead.

\subsection{A new efficient approximate convex hull algorithm in high dimensions}\label{subsec-algorithm}

In this section, we propose our algorithm for finding the convex hull of high dimensional points, which is called \textbf{Revised Greedy Expansion (RevisedGE)}, by taking \textbf{Greedy Expansion (GE)} \cite{HosseinACH2016} as our reference.
As what we mentioned above, characterizing the exact convex hull is of high computation complexity.
Hence, our algorithm is to provide an approximate convex hull with a balance between complexity and accuracy.
Instead of characterizing a polyhedron by its facets as what the usual way does, we give the out vertices of our approximate convex hull.
In other words, for a given set of points $\mathcal{S}$, we aim to select some points from $\mathcal{S}$ to compose the approximate convex hull.

To find a convex hull, the most critical problem is how to judge whether a point $\mathbf{v}$ is inside or outside a set of points $\mathcal{S}=\{\mathbf{x}_1,\cdots,\mathbf{x}_S\}$.
We achieve it by calculating the distance from $\mathbf{v}$ to $\mathcal{S}$.
The distance $d(\mathbf{v}, \mathcal{S})$ is defined as the following dimension independent quadratic programming:
\begin{equation}
\label{eq-qp}
\begin{aligned}
	d(\mathbf{v}, \mathcal{S}) =	& \min_{\alpha_i} \quad ||\mathbf{v} - \sum_{i=1}^S{\alpha_i \mathbf{x}_i}||_2 \\
									& s.t. \quad \alpha \geq 0, \sum_{i=1}^{S}\alpha_i = 1.
\end{aligned}
\end{equation}

This quadratic programming can give us a point $\hat{\mathbf{x}}$, being interior of the convex hull of $\mathcal{S}$, which is closest to the given point $\mathbf{v}$.
The distance from point $\mathbf{v}$ to the set $\mathcal{S}$ is defined as the distance from point $\mathbf{v}$ to point $\hat{\mathbf{x}}$.
Therefore, if a point $\mathbf{v}$ is already an interior of the convex hull of $\mathcal{S}$, the corresponding $\hat{\mathbf{x}}$ will be itself and then $d(\mathbf{v}, \mathcal{S})=0$.

Moreover, with the distance, for a given set of points $\mathcal{S}$, we formulate the $\epsilon$-approximation convex hull $\mathcal{E}$ by requiring
 $$ \max_{\mathbf{v} \in \mathcal{S}} d(\mathbf{v},\mathcal{E}) \leq \epsilon.$$

Then we adopt a greedy expansion method to select points from $\mathcal{S}$ to compose the $\epsilon$-approximation convex hull.
For each iteration, supposing $\mathcal{E}$ is the set of selected points, we will find
 $$ \mathbf{x}=\arg\min\limits_{\mathbf{x}\in\mathcal{S}\backslash\mathcal{E}}\max\limits_{\mathbf{v}\in\mathcal{S}\backslash\mathcal{E}}d(\mathbf{v}, \mathcal{E}\cup\mathbf{x}) $$
and add it to $\mathcal{E}$.

Note that the time complexity for finding such $\mathbf{x}$ is $O((S-E)^2)$, which is extremely high when the size of $\mathcal{E}$ is small.
Based on this point, we apply \textbf{Kernelized Convex Hull Approximation (KCHA)} method \cite{HuangKCH2018} to obtain the startup set.
This method is based on Semi-Nonnegative Matrix Factorization (Semi-NMF), in which the kernel trick can be utilized to speed up finding the extreme points.
The full algorithm is presented as Algorithm \ref{alg-ach}.

\begin{algorithm}[t]
\small
\caption{Revised Greedy Expansion Algorithm}
\label{alg-ach}
\KwIn{points $\mathcal{S}=\{\mathbf{x}_1,\mathbf{x}_2,\cdots,\mathbf{x}_S\}$, approximation rate $\epsilon$.}
\KwOut{$\epsilon$-approximation convex hull $\mathcal{E}$.}

Find kernelized extreme points through \textcolor{black}{KCHA} and initialize $\mathcal{E}$ as the set of these points\;
Assign the set of outside points $\mathcal{R}=\{\mathbf{x} \in \mathcal{S}|d(\mathbf{x}, \mathcal{E})>0\}$\;
\While{$\max_{\mathbf{x} \in \mathcal{S}} d(\mathbf{x},\mathcal{E}) > \epsilon$}{
	Select point $\mathbf{x}=\arg\min\limits_{\mathbf{x}\in\mathcal{R}}\max\limits_{\mathbf{v}\in\mathcal{R}}d(\mathbf{v}, \mathcal{E}\cup\mathbf{x})$ and add it to $\mathcal{E}$\;
	\For{point $\mathbf{x}$ in $\mathcal{E}$}{
		\If{$d(\mathbf{x}, \mathcal{E}\backslash\mathbf{x})==0$}{
			Remove point $\mathbf{x}$ from $\mathcal{E}$\;
		}
	}
	\For{point $\mathbf{x}$ in $\mathcal{R}$}{
		\If{$d(\mathbf{x}, \mathcal{E})==0$}{
			Remove point $\mathbf{x}$ from $\mathcal{R}$\;
		}
	}
}
\Return $\epsilon$-approximation convex hull $\mathcal{E}$\;
\end{algorithm}

\begin{figure}[t]
	\includegraphics[width=\linewidth]{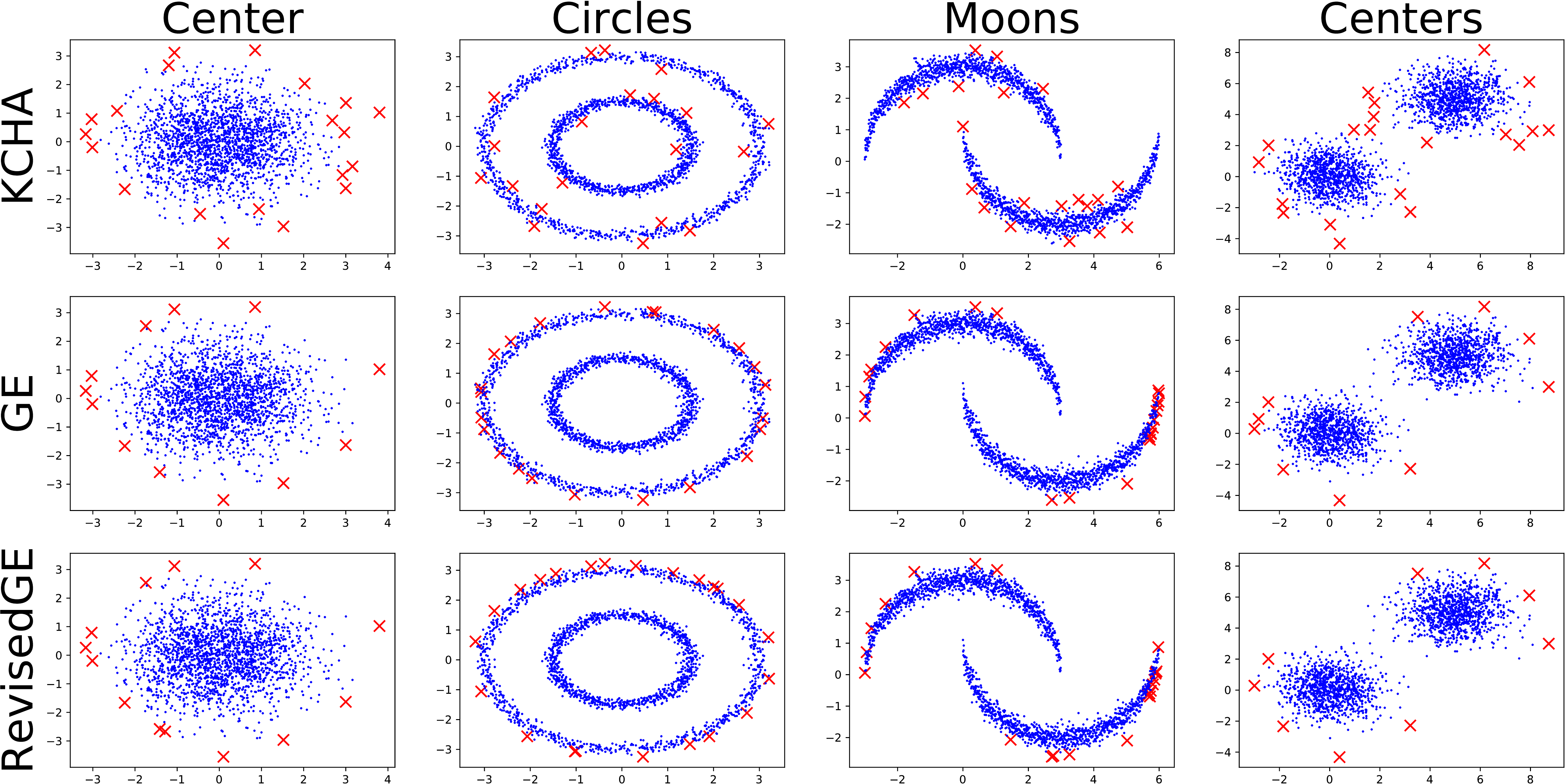}
	\caption{Approximate Convex Hull found by the three algorithms on four toy datasets. Both blue nodes and red crosses are the points inputted into the algorithms, and red crosses represent points selected as extreme points by the respective algorithm.}
	\label{fig-approximation_convex_hull}
\end{figure}

In summary, GE is dimension-independent, yet the expanding of convex hull $\mathcal{E}$ is quite time consuming, especially for the early stage of the expansion.
On the other hand, though KCHA gives a relatively fast algorithm, experiments show that it is relatively less accurate.
By the combination of GE and KCHA, RevisedGE achieves both the accuracy of GE and benefits in the high efficiency of KCHA.
To illustrate the difference among them, we generate four 2-dimension toy datasets and apply the three algorithms on them.
Figure \ref{fig-approximation_convex_hull} shows the results of the three algorithms.
As we can see, both GE and RevisedGE outperform KCHA on accuracy since KCHA select many interior points.
Moreover, we evaluate the efficiency of the three algorithms by directly comparing their running time, as shown in Table \ref{tab-running_time}.
Benefiting from obtaining the startup convex hull by KCHA, compared to GE, RevisedGE achieves huge improvement on efficiency.

\subsection{Experiment platform}\label{subsec-platform}

As mentioned in \S\ref{sec-intro}, this paper focuses on the study of fully connected layers.
Four representative datasets of two categories are used.
Specifically, the first category contains two well-studied image classification datasets.
\begin{description}
   \item[MNIST \cite{mnist}:] the well-known database of handwritten digits, which contains 65,000 (55,000 for training and 10,000 for test) images of 10 classes.
   \item[CIFAR-10 \cite{cifar-10}:] a collection of images that are widely used to train computer vision algorithms, which contains 60,000 (50,000 for training and 10,000 for test) color images in 10 classes.
\end{description}
The second category contains two datasets for the graph representation learning task extracted from AceKG \cite{acekg}, a large-scale knowledge graph in academic domain:
\begin{description}
   \item[FOS-CS-5:] a network containing all the papers, authors and venues in AceKG under 5 subfields of Computer Science\footnote{\scriptsize{5 subfields: Computer Security, Operating System, Algorithm, World Wide Web and Machine Learning}}, which contains 6,465,137 nodes in 5 classes.
   \item[GOOGLE-8:] with matching 8 categories of venues in Google Scholar\footnote{\scriptsize{\url{https://scholar.google.com/citations?view\_op=top\_venues\&hl=en\&vq=eng}}} to those in AceKG, a heterogeneous network containing all the papers published on 151 of 160 venues (8 categories $\times$ 20 per category) and their authors, which contains 1,236,127 nodes in 8 classes.
\end{description}
More details about the two datasets refer to \cite{acekg}.
In this paper, to feed such network dataset into the neural network, we adopt the well-known graph representation learning algorithm DeepWalk \cite{deepwalk} to embed both the two networks into 64-dimension vectors and randomly select 60,000 (50,000 for training and 10,000 for test) vectors for the following study.
Graph representation learning, also known as network embedding, has been consistently studied in recent years and benefits most network analysis tasks including link prediction \cite{wang2018shine}, recommendation \cite{GraphGAN} and community detection \cite{CommunityGAN}.
It aims to learn graph structure and encode each node in the graph into a low-dimensional vector, which can be fed into a wide range of machine learning algorithms, including the neural network considered in this paper.
The two embedded vector datasets are available online\footnote{\scriptsize{http://bit.ly/IJCAI2019-Submission-367}}.

\begin{table}[t]
\centering
\begin{tabular}{l||r|r|r|r}
Algorithm & Center  & Circles & Moons   & Centers  \\
\hline
\hline
KCHA      & 7.78    & 16.15   & 18.19   & 18.06    \\
GE        & 2074.81 & 2855.46 & 6655.60 & 7568.68  \\
RevisedGE & 9.13    & 32.63   & 38.88   & 45.73
\end{tabular}
\caption{Running Time (Sec.) of the three algorithms on four toy datasets.}
\label{tab-running_time}
\end{table}

There is a distinct difference among these datasets:
though all of them can be seen as vectors, the images in MNIST and CIFAR-10 are originally continuous dense vector, while the other two datasets are the embedding for graph structure.
As our work concentrates on the study of vectors, we believe it would be more persuasive to consider both of them.

As a beginning, we train a neural network for each dataset.
Here all the network structures are the same, which contain four fully connected layers and 1024 neurons per layer.
The detailed statistics for the well trained neural networks are shown in Table \ref{tab-networkstatistics}.
The accuracies of the networks on training set are pretty high, where the lowest accuracy (about 80\% on FOS-CS-5) is still acceptable considering the difficulty for the 5-class classification problem.
Even though the networks on CIFAR-10 and FOS-CS-5 seem to be overfitting, they still show the ability for correctly classifying the data points in the training set.
So we think it is meaningful to study the geometric structure of their activation spaces.

\begin{table}[t]
\centering
\begin{tabular}{l||r|r|r|r}
Dataset      & $\ell$ & $d$  & TR    & TE   \\
\hline
\hline
MNIST        & 4   & 1024 & 100\%   & 94.8\% \\
CIFAR-10     & 4   & 1024 & 94.7\%  & 46.6\% \\
FOS-CS-5     & 4   & 1024 & 79.7\%  & 59.4\% \\
GOOGLE-8     & 4   & 1024 & 99.2\%  & 95.1\%
\end{tabular}
\caption{Neural network statistics. $\ell$: number of layers, $d$: number of neurons per layer, TR: Accuracy on the training set, TE: Accuracy on the test set.}
\label{tab-networkstatistics}
\end{table}

We then apply the RevisedGE algorithm on the activation spaces of the models listed in Table \ref{tab-networkstatistics}.
The main results of several related studies are summarized in the next section.

\section{Geometric Structure of Activation Spaces}\label{sec-STRU}

In this part, we build a geometric description for the convex hulls of data of different classes in the activation spaces.
We believe such approximation gives us a better understanding of the intrinsic property of the neural networks and our main results do expose some surprising phenomena.
Here we highlight the four most significant results, which show a rather clear picture of the macrostructure of activation spaces.

For a well-trained neural network $\network{N}$ with $\ell$ fully connected layers, where the $i$th layer contains $d_i$ neurons, when any one data point $\mathbf{x}$ from the whole dataset $\mathcal{X}$ is inputted to $\network{N}$, the output at the $i$th layer can be regarded as a vector of dimension $d_i$.
We use the symbol $\mathbf{x}_i \in \mathbb{R}^{d_i}$ to represent this vector and call it the {activation vector} of input $\mathbf{x}$ at layer $i$.
We call the set of all activation vectors in each layer the activation space of that layer.
For each layer, our goal is to figure out the geometrical shape formed by activation vectors of different classes.
We will highlight the class information by the superscript.
For example, for a certain class $C$, we denote $\mathbf{x}_i^C$ as activation vector $\mathbf{x}_i$ (at the $i$th layer) of the input $\mathbf{x}$ of class $C$.
When we are discussing a set of data, we use $\mathcal{Y}_i^{C}=\{\mathbf{y}_{i,1}^C, \mathbf{y}_{i,2}^C,\ldots, \mathbf{y}_{i,Y}^C\}$ to stand for a collection of $Y$ activation vectors of the same class in the $i$th layer.
For convention, we will use $\ch{\mathcal{X}_i}$ to stand for the convex hull we get by implementing Algorithm \ref{alg-ach} on dataset $\mathcal{X}$ in layer $i$.
Further on, $\ach_{i}^{C}$ is adopted as an abbreviation for $\ch{\mathcal{Y}_i^{C}}$.
Note sometimes we may drop the class and/or layer indexes for simplicity when they are evident from the context.

\subsection{All points are crucial}\label{subsec-apc}
For any activation vector $\mathbf{y}_{i,k}^{C}$, it is obvious that if $\mathbf{y}_{i,k}^{C}\in \ch{\mathcal{Y}_i^{C} \backslash \mathbf{y}_{i,k}^{C}}$, then $\mathbf{y}_{i,k}^{C}$ is not an extreme point, which means it is redundant from the perspective of constructing the convex hull.
Though the famous high-dimensional curse shows that most of the data would lie on the surface of a high-dimensional sphere, it does not give information on the necessity of them.
Actually, we thought that there should be activation vectors that lie in the inner part of the convex hull.

Surprisingly, for all the datasets, we implement the RevisedGE algorithm to get $n$ convex hulls (according to the $n$ classes in the dataset) at each layer for the training and test datasets.
For every data point $\mathbf{x}$ with class $C$ and its corresponding active vector $\mathbf{x}_i$ in the $i$th layer ($1\leq i\leq 4$), the algorithm shows that $\mathbf{x}_i$ always turns out to be an extreme point in $\ach_{i}^{C}$.
In other words, no vector $x_i$ can be represented as a convex combination of the remaining vectors of the same class.
Mathematically, our results show that for any class $C$, any layer $i$ and any dataset $\mathcal{Y}^{C}$ in consideration:
$$ \mathbf{y}_{i,k}^C\not\in \ach(\mathcal{Y}_i^{C}\setminus \mathbf{y}_{i,k}^C). $$

Intuitively speaking, if we consider the convex hull as an approximation of the knowledge learnt by the neural networks, this result shows that all vectors are essential to the corresponding conceptions.
It means that the natural figures in the original dataset, even if they are of the same class, are almost surely independent from each other from the view of the fully connected neural networks.
One possible explanation for this phenomenon is that, in our settings, the neural networks learn something different from every input.

\subsection{No mis-inclusion}\label{subsec-misin}
The results in \S\ref{subsec-apc} shows that no activation vector is covered by the convex hull of the same class without that vector.
Then what if we consider the relationship between an activation vector and convex hulls of different classes?
To answer this question, for all four layers, we calculate the distance according to the definition in Eq. (\ref{eq-qp}) for all possible combinations of nodes and convex hulls, which shows
$$ \mathbf{y}_{i,k}^{C_1}\not\in \ach_i ^{{C_2}} \text{~~given~~} {C_1}\neq {C_2},$$
i.e., none of the active vectors from the training or test datasets lies inside the convex hull of a different class.

There are two more remarks on this phenomenon:
\textbf{1)} So far this statement can be rigorously proved only for the output layer for the convexity of the Softmax function.
We thought earlier that the lower layers could be relatively more `confused' and might sandwich the vectors of different classes.
Surprisingly, as far as the neural networks we trained are considered, it works pretty well in the sense of no mis-inclusion.
\textbf{2)} One may argue that a natural question considering convex polygons would be whether they are separate from each other.
Unfortunately, there is an infeasible computational lower bound for this question.

\subsection{Consistent spacial distribution of activation vectors of the same class}\label{subsec-spacial}
\begin{figure}[t]
	\centering
	\includegraphics[width=\linewidth]{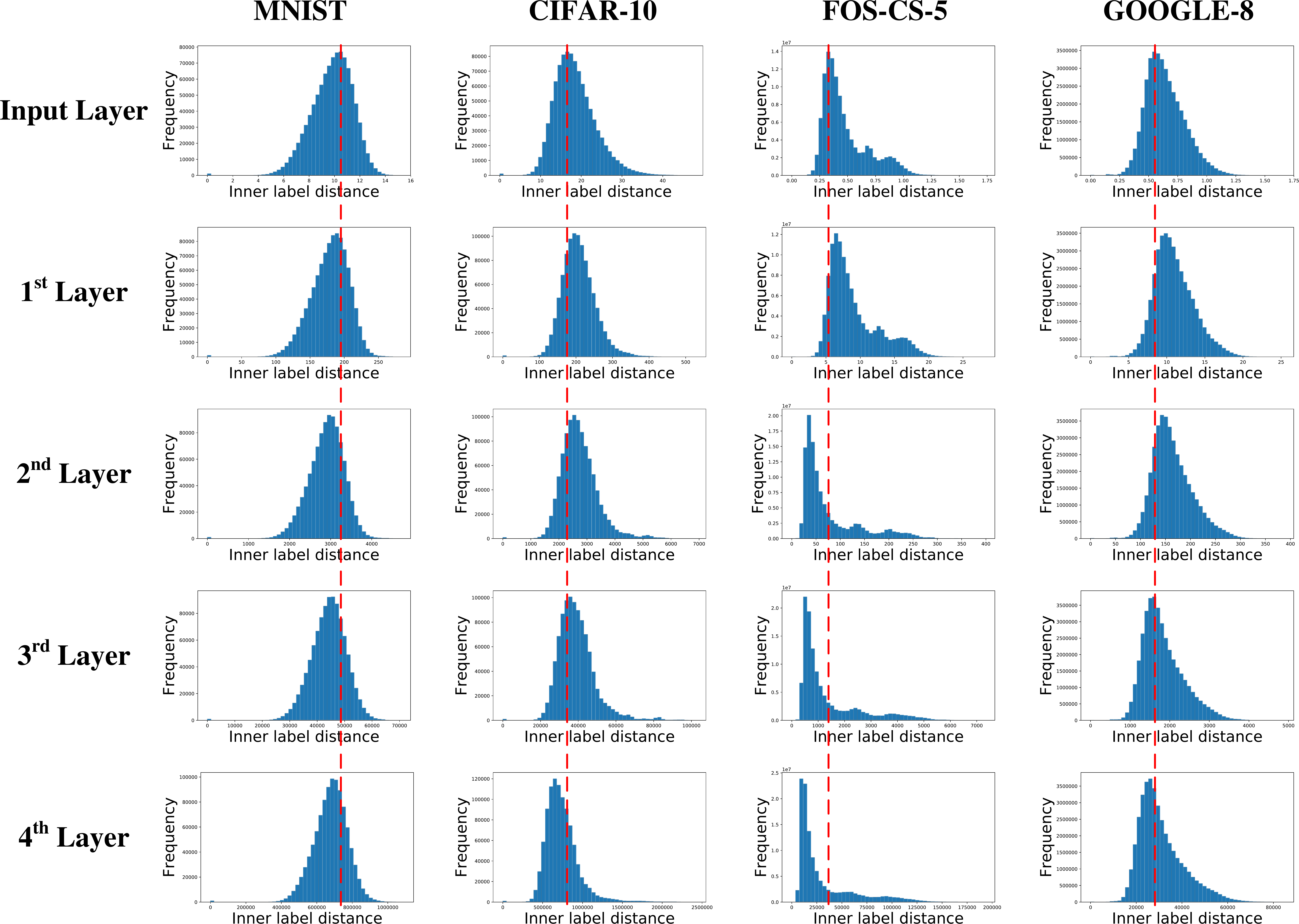}
	\caption{Frequency distribution histogram of inner class distance for five layers on four datasets. Each column represents one dataset and rows from upper to lower illustrate the change of the distribution of inner class distance from input layer (original data) to four hidden FC layers. The red dash line marks the peak of the input layer, helping to observe the shift of the peak through layers.}
	\label{fig-inner_dis}
\end{figure}

Besides the inclusion relation between vectors and convex hulls, another critical problem is the spatial distribution of activation vectors (of the same class).
Here we try to solve this question by calculating the Euclidean distances between any two active vectors of the same class.
In another word, for any layer $i$ and class $C$, we calculate
 $$d(\mathbf{y}_{i,k_1}^C, \mathbf{y}_{i,k_2}^C) = ||{\mathbf{y}_{i,k_1}^C - \mathbf{y}_{i,k_2}^C}||_2 \text{~~where~~} {k_1}\neq {k_2}.$$
We visualize the results by drawing the distribution histogram of the distance for each layer and each class.
For the space limitation, here we only present the result of one class of data for each dataset, which is shown in Figure \ref{fig-inner_dis}.
It should be mentioned that the results of the remaining classes of data are very similar.


As we can see, in general, the distribution of the inner class distances is unimodal.
It means that activation vectors of the same class distribute almost uniformly and thus there would not be many small clusters which lie distant from each other.
Moreover, for deeper layers, i.e., closer to the output layer, there are two apparent phenomena:
\textbf{1)} The absolute value of the inner class distance increases (abscissa increases almost exponentially).
Actually, this is closely related to the existence of adversarial examples \cite{goodfellow2014explaining}.
\textbf{2)} Through the four hidden layers, overall, the peaks of the distribution gradually shift to the left, which implies that activation vectors of the same class tend to (though in different metric) get closer to each other.


\subsection{Monotone tendency of inter class distance}

Besides the inner class distance in \S\ref{subsec-spacial}, we give a further study on the inter class distance, i.e., Euclidean distance between two vectors with different classes.
We use the mean of the distance as the measure for this study.
We define

 $$d'(\mathcal{Y}_i^{C_1}, \mathcal{Y}_i^{C_2}) = \sum_{k_1=1}^{Y_i^{C_1}}\sum_{k_2=1}^{Y_i^{C_2}}\frac{||{(\mathbf{y}_{i,k_1}^{C_1}- \mathbf{y}_{i,k_2}^{C_2})}||_2}{ Y_i^{C_1}Y_i^{C_2}}~~\text{for}~~ C_1\neq C_2.$$

\begin{figure}[t]
	\centering
	\subfigure[]{\label{fig-inter_dis-a}\includegraphics[width=0.435\linewidth]{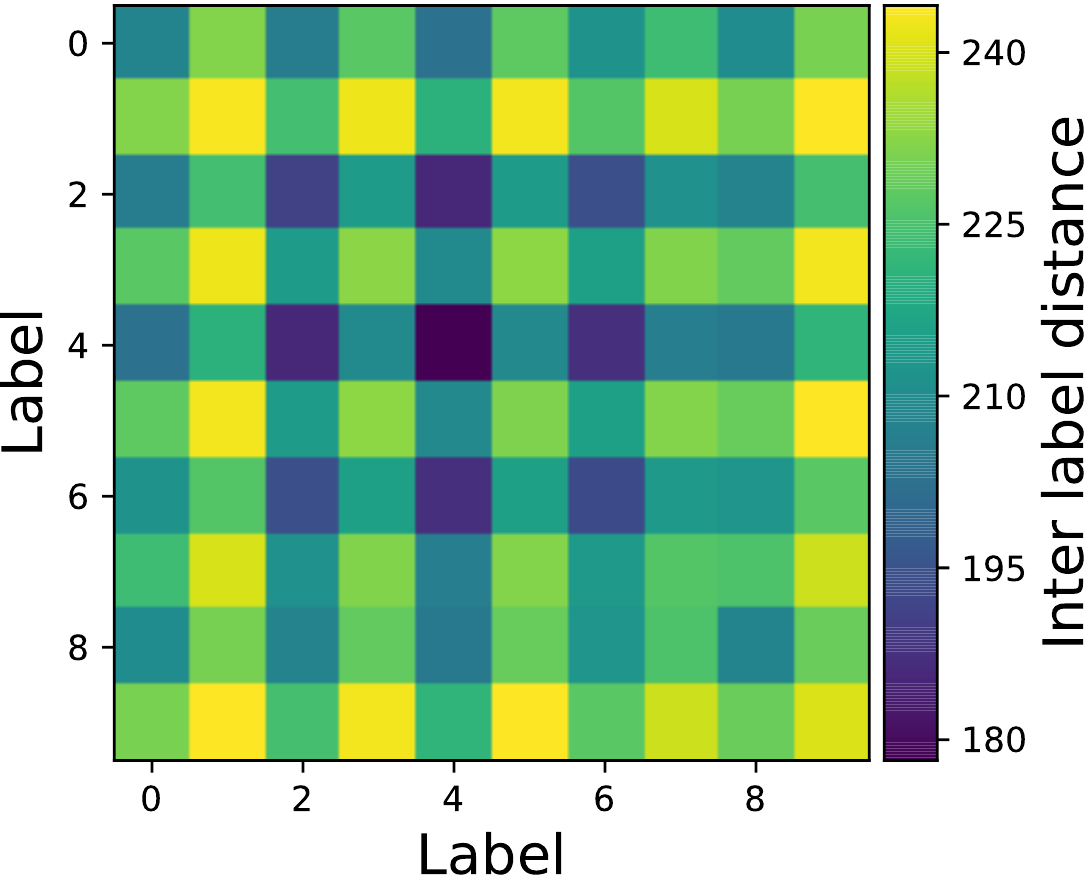}}
	\subfigure[]{\label{fig-inter_dis-b}\includegraphics[width=0.520\linewidth]{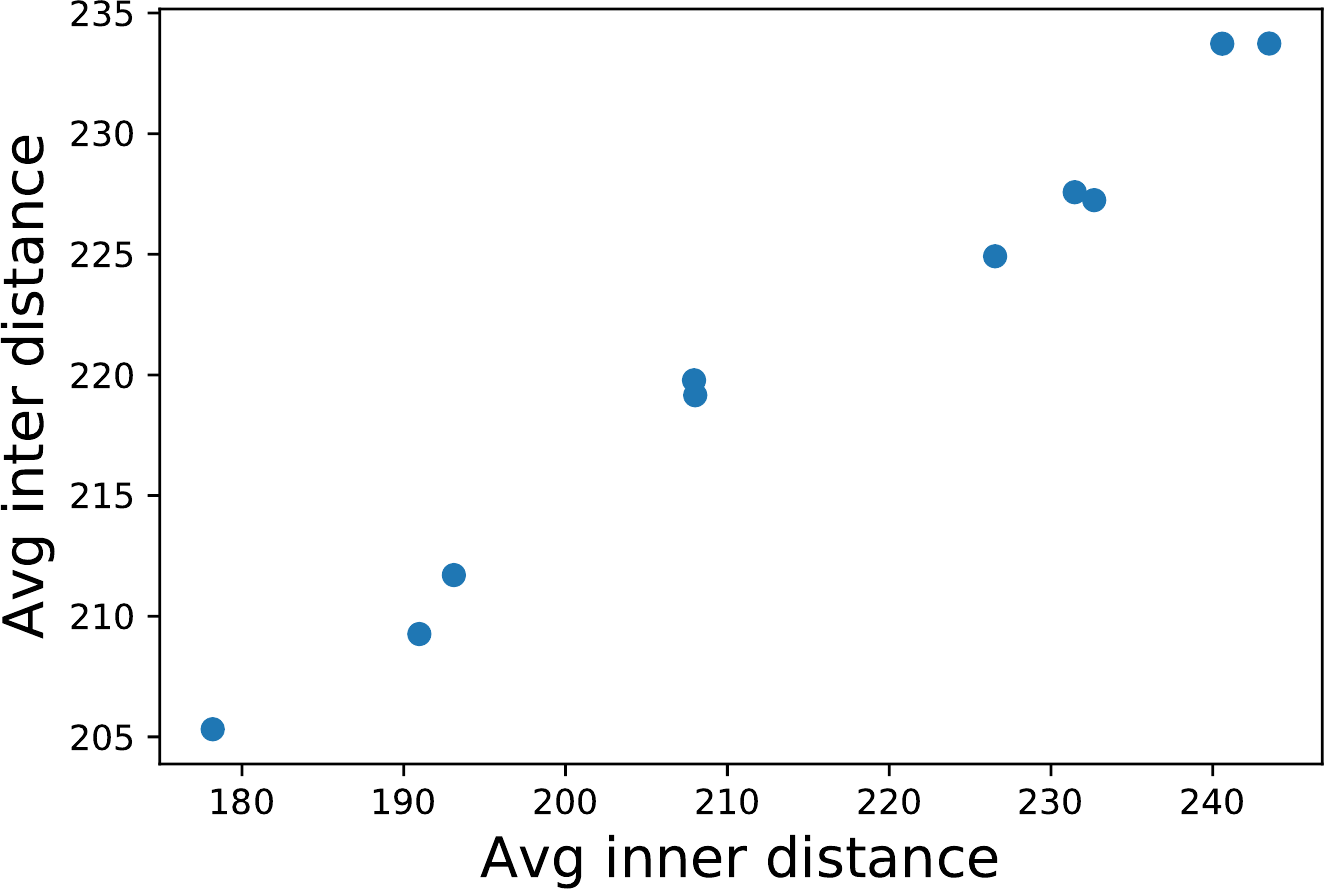}}
	\label{fig-inter_dis}
	\caption{(a) Distribution of inter class distance, where each ($C_1,C_2$) element means the distance from class $C_1$ to $C_2$. (b) Relationship between inner and inter class distances.}
\end{figure}

For space limitation, we only take the inter class distance of the first layer on CIFAR-10 as an example, which is shown in Figure \ref{fig-inter_dis-a}.
In this figure, each ($C_1,C_2$) element stands for the distance from class $C_1$ to $C_2$.
For example, the first row shows the $d'$ distance from class $0$ to all $10$ classes.
Note that the diagonal of the matrix represents the average inner distance of the $10$ classes.

The most interesting phenomenon in this part is that, for a class of data with larger inner class distance increase, the distance of that class of data to data of other classes also gets larger.
For example, class $1$ has the largest inner distance, then the distance from class $1$ to other classes are also relatively larger.
For a more intuitive understanding,  as shown in Figure \ref{fig-inter_dis-b}, we draw the relationship between inner and inter class distances for the first layer on CIFAR-10.
We can conclude that the inner and inter distances are positively related to each other in general.
To figure out the reason and meaning of this phenomenon is an interesting future work.

\subsection{Hypothesis on the structure of activation spaces}

So far we have given our main four results on the geometric structure of activation spaces.
Here we give our hypothesis on the structure of activation spaces:
\begin{quote}
\noindent{\bf{Hypothesis}.} The groups of activation vectors distribute nonuniformly on a high-dimensional concentric sphere. Datasets of different classes lie in different orbits, with the greater the radius, the larger the inner distance.
\end{quote}

Before the end of this section, we would like to include one further discussion about the above hypothesis.
As we have discussed in \S\ref{sec-actSpace}, activation vectors are data in high-dimensional space.
Though we have tried to characterize the data via convex hull and inner/inter class distance, one natural question is the volume of the convex hull and their intersection.
Note that though our experiments show that no data is covered by the convex hulls of different classes, it does not imply that different classes of convex hull are disjoint.
Simple counterexamples  can be visualised in 2-dimensional space.
Volume information, besides its geometrical importance, can give a straightforward answer to the distribution of different classes of training data in activation space.
However, this problem is computationally hard in high dimensional space.
There have been a series of random polynomial algorithms for approximating the volume of a convex body.
They all assume the existence of a membership oracle, however, even with such assumption the classic algorithms like \cite{DFK89stoc,lovasz2003focs} are not practically feasible when we come to the settings of hundreds of thousands dimension.
The computation of the volume of the intersection of high-dimensional geometric objects has been proved to be $\mathbf{\#P}$-hard \cite{Bringmann2010} even for boxes.

\section{Nearest Convex Hull vs. Well Trained Neural Networks}\label{sec-CCH}

In this section, we deepen our work by discussing the benefits of studying the geometric structure of activation spaces in the sense of classification performance.
Based on the observation in \S\ref{sec-STRU}, we apply the so-called \emph{nearest convex hull classification algorithm} on activation vectors of every layer.
The outputs are then used as a new classification decision and count the accuracy.
This new accuracy could be regarded as a new metric for evaluating the functionality of each layer.

\subsection{The nearest convex hull classification algorithm}

Similar to the traditional strategy in neural networks, we divide the datasets into training and test sets.
For each layer in the trained networks, denoted as the $i$th layer, we run the nearest convex hull classification algorithm as follows:
\begin{itemize}
\item For any activation vector $\mathbf{x}_i$ in training set, take $n$ convex hulls formed by the activation vectors of all training data except $\mathbf{x}_i$, where $n$ is the number of possible classes in the considered model.
Different convex hulls correspond to different classes, and $\mathbf{x}_i$ will take the class of the convex hull to which it is closest.
Further on, the training accuracy is defined as the percent of vectors whose predicted class is the same as its ground truth.
\item For the test set data, we label all the activation vectors of the $i$th layer as above, and the same goes to test accuracy.
Only now the convex hulls are formed by the whole training set.
\end{itemize}

One big difference to the common neural networks is that now we can calculate the training/test accuracy for all layers, rather than only consider the output of the last layer.
We run the nearest convex hull classification algorithm on all four datasets introduced in \S\ref{subsec-platform}.
The experiment results are summarized in Table \ref{tab-nearestconvexhull}.

\begin{table}[t]
\small
\centering
\begin{tabular}{|l|l|l|l|l|l|}
\hline
                          &       & \multicolumn{1}{c|}{\begin{tabular}[c]{@{}c@{}}$1$st\\Layer\end{tabular}} & \multicolumn{1}{c|}{\begin{tabular}[c]{@{}c@{}}$2$nd\\ Layer\end{tabular}} & \multicolumn{1}{c|}{\begin{tabular}[c]{@{}c@{}}$3$rd\\ Layer\end{tabular}} & \multicolumn{1}{c|}{\begin{tabular}[c]{@{}c@{}}$4$th\\ Layer\end{tabular}} \\ \hline\hline
\multirow{2}{*}{MNIST}      & train & 0.988 & 0.987 & 0.986 & 0.984 \\ \cline{2-6}
                            & test  & 0.983 & 0.980 & 0.979 & 0.976 \\ \hline\hline
\multirow{2}{*}{CIFAR-10}   & train & 0.564 & 0.583 & 0.611 & 0.623 \\ \cline{2-6}
                            & test  & 0.522 & 0.519 & 0.508 & 0.494 \\ \hline\hline
\multirow{2}{*}{FOS-CS-5}      & train & 0.738 & 0.731 & 0.730 & 0.730 \\ \cline{2-6}
                            & test  & 0.678 & 0.675 & 0.672 & 0.666 \\ \hline\hline
\multirow{2}{*}{GOOGLE-8} & train & 0.963 & 0.969 & 0.977 & 0.979 \\ \cline{2-6}
                            & test  & 0.939 & 0.948 & 0.954 & 0.955 \\ \hline
\end{tabular}

\caption{Accuracy of nearest convex hull classification for activation space of each layer.}
\label{tab-nearestconvexhull}
\end{table}

\begin{figure}
	\centering
	\includegraphics[width=\linewidth]{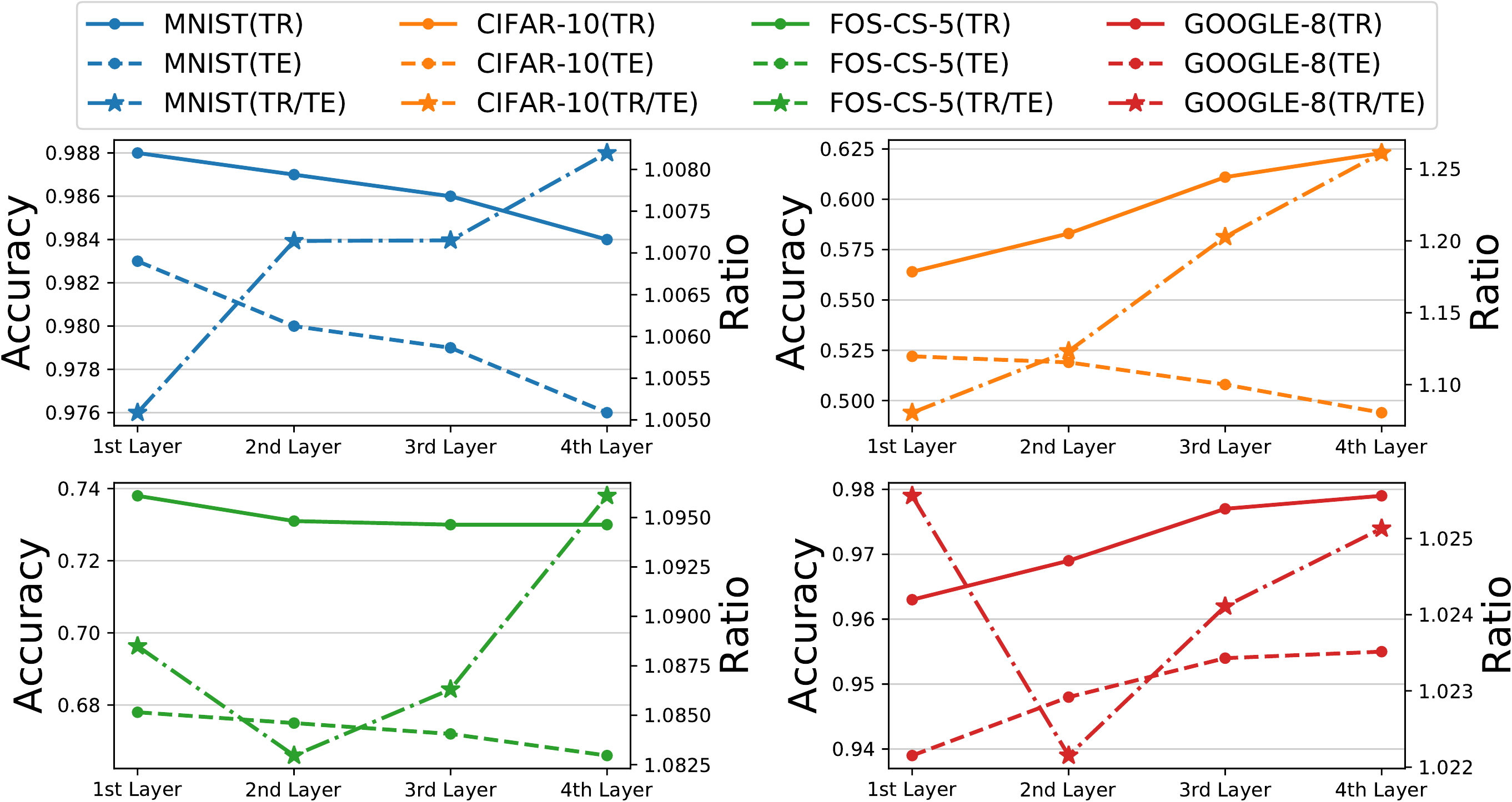}
	\caption{Accuracy of nearest convex hull classification for activation space of each layer on four datasets. `TR' and `TE' represent training accuracy and test accuracy respectively.}
	\label{fig-nearestconvexhull}
\end{figure}

Comparing to the results in Table \ref{tab-networkstatistics}, the nearest convex hull classification method performs better in every test sets (up to more than 7\% better) in all four layers.
In our opinion, this is a very interesting phenomenon, of which three key points should be addressed here
\begin{enumerate}
	\item Intuitively nearest convex hull classification is less involved than the neural networks, for the former has a rather clear geometric meaning.
	The better accuracy could be seen as a proof of the soundness of the results in \S\ref{sec-STRU}: there does exist a relatively simple geometric structure which can be utilized for classification.
	\item Now take both the training accuracy and test accuracy into account.
	Table \ref{tab-networkstatistics} shows that, for the well trained neural networks, though accuracies of the four networks on training set are pretty high, there are significant accuracy gaps between training sets and test sets. For example, in  CIFAR-10 the gap is 48.1\%, and in FOS-CS-5, 20.3\%.
	Thus there exists serious overfitting problem in these settings.
	However, when we take the nearest convex hull classification algorithm, the corresponding gaps are just 12.9\% and 6.4\% respectively.
	It opens a new perspective on the study of overfitting/underfitting.
	We believe the nearest convex hull classification provides a more reliable accuracy metric in the training process which can be used to evaluate the quality of the models.

	\item As being visualized in Figure \ref{fig-nearestconvexhull}, there is manifest monotonicity considering the accuracy statistics of the nearest convex hull classification algorithm.
	Especially for the MNIST and CIFAR-10 data, the ratio of training accuracy over test accuracy is monotone increasing, meanwhile, the layers while the test accuracy is monotone decreasing.
	It implies that if we take the nearest convex hull classification algorithm, then it will be better to just keep the output in the first activation space.
	The rest of the neural network can be simply removed.
\end{enumerate}

\subsection{Comparison with classical clustering algorithms}

One natural question about the nearest convex hull classification algorithm on the activation spaces is can we replace it by simpler clustering algorithms like $k-$nearest neighbors.
For a better understanding of the four datasets and the trained neural networks, we also apply some benchmark algorithms on the four datasets.
As shown in Table \ref{tab-benchmarks}, logistic regression (LR), support vector machine (SVM) and $k-$nearest neighbors (KNN) are adopted, whose training and test accuracy on the four datasets are listed.

\begin{table}[t]
\centering
\small
\begin{tabular}{|l|l|c|c|c|}
\hline
                          &         &  LR   &  SVM  &  KNN  \\ \hline\hline
\multirow{2}{*}{MNIST}      & train & 0.928 & 0.942 & 0.981 \\ \cline{2-5}
                            & test  & 0.920 & 0.944 & 0.968 \\ \hline\hline
\multirow{2}{*}{CIFAR-10}   & train & 0.495 & 0.445 & 0.505 \\ \cline{2-5}
                            & test  & 0.388 & 0.440 & 0.340 \\ \hline\hline
\multirow{2}{*}{FOS-CS-5}   & train & 0.589 & 0.547 & 0.747 \\ \cline{2-5}
                            & test  & 0.587 & 0.546 & 0.638 \\ \hline\hline
\multirow{2}{*}{GOOGLE-8}   & train & 0.944 & 0.941 & 0.966 \\ \cline{2-5}
                            & test  & 0.940 & 0.938 & 0.955 \\ \hline

\end{tabular}
\caption{Result of benchmarks on the four datasets. LR: Logistic Regression, SVM: Support Vector Machine, KNN: K-Nearest Neighbors.}
\label{tab-benchmarks}
\end{table}

One can see that nearest convex hull classification algorithm outperforms the classic algorithms on MNIST, CIFAR-10 and FOS-CS-5 datasets and almost level the score on GOOGLE-8.
It means, comparing to the other strategy, convex hull does reveal more information about the intrinsic property of the activation vectors.

\section{Related Work}\label{sec-related}

In addition to the comparison presented in \S\ref{sec-intro}, our work has a tight connection with some of the earlier work on trying to demystify the deep neural networks.
In \cite{Raghu2017Expressive}, the authors have found that the complexity of the computed function grows exponentially with the depth, and that trained networks are far more sensitive to their lower layer weights: they are much less robust to noise in these layer weights.
Surely convex hull can be regarded as a special kind of trajectory defined in their paper which has strong intuitive meanings: the extreme points are those conceptions which cannot be got by the linear combination of others.
Our results on MNIST data correspond to the study in \cite{Raghu2017Expressive} where they find that training lower hidden layers can lead to better performance.
Comparatively, their work gives the phenomenon while our work gives a geometric explanation to this phenomenon: the lower layers do matter more in the sense they actually produce higher accuracy according to the metric of the closest convex hull.

Last but not least, as being shown in our work on the approximate convex hull of activation vectors and its relating application in neural network optimization, it is reasonable to take convexity into account.
However, (non-)convexity is a fundamental important factor in the study of generalization error \cite{Jordan2017} \cite{Jianli2019Archive}.
Generalization guarantee becomes difficult to achieve when it comes to highly non-convex settings like deep neural networks.
We think the geometric structure given in our work shed light on this problem: as the nearest convex hull algorithm works well, it is reasonable to take less non-convexity assumption on neural networks and this should simplify the problem.

\section{Conclusion} \label{sec-conclusion}
Our work tries to bridge the gap between the big success of modern neural networks and the much less understanding of why they work so well.
Comparing to the previous efforts on this topic, we propose a simple and neat idea, i.e., to study the convex hull of corresponding conceptions formed by every layer of the neural networks.
In order to fulfill this goal, we build a new approximation algorithm which works efficiently in the high dimensional situation.
Several surprising results on the functionality of neural networks are given in this paper, especially that
1) the neural networks are quite smart in the sense that there is no mis-inclusion in the sense of convex hull;
2) the closest convex hull algorithm outperforms the deep neural networks even from the lower hidden layers.
As the convex hull has explicit geometric meanings compared to the deep neural networks,  we believe our work shed light on the intrinsic properties of neural networks.
There are two important future areas we want to explore thoroughly.
One is to take the convolutional neural network into consideration, and the other one is to combine the convex hull into the optimization or training of neural networks.

\section*{Acknowledgements}
Special thanks should be given to Prof. John E. Hopcroft who brings this interesting topic to us and gives us  many valuable suggestions.

This work was financially supported by National Key Research and Development Program of China (Grant No. 2017YFB0701900) and NSFC (Grant No. 61572318, 61772336, 61872142).

\newpage

\bibliographystyle{named}

\begin{thebibliography}{}

\bibitem[\protect\citeauthoryear{Bringmann and Friedrich}{2010}]{Bringmann2010}
Karl Bringmann and Tobias Friedrich.
\newblock Approximating the volume of unions and intersections of
  high-dimensional geometric objects.
\newblock {\em Computational Geometry}, 43(6-7):601--610, 2010.

\bibitem[\protect\citeauthoryear{Chand and Kapur}{1970}]{Donald1970}
Donald~R Chand and Sham~S Kapur.
\newblock An algorithm for convex polytopes.
\newblock {\em Journal of the ACM (JACM)}, 17(1):78--86, 1970.

\bibitem[\protect\citeauthoryear{Dwyer}{1988}]{DwyerPhD1988}
Rex~Allen Dwyer.
\newblock {\em Average-case Analysis of Algorithm for Convex Hulls and Voronoi
  Diagrams.}
\newblock PhD thesis, CMU, USA, 1988.

\bibitem[\protect\citeauthoryear{Dyer \bgroup \em et al.\egroup
  }{1988}]{DFK89stoc}
Martin Dyer, Alan Frieze, and Ravindran Kannan.
\newblock {\em A random polynomial time algorithm for estimating volumes of
  convex bodies}.
\newblock University of Leeds, School of Computer Studies, 1988.

\bibitem[\protect\citeauthoryear{Goodfellow \bgroup \em et al.\egroup
  }{2014}]{goodfellow2014explaining}
Ian~J Goodfellow, Jonathon Shlens, and Christian Szegedy.
\newblock Explaining and harnessing adversarial examples.
\newblock {\em arXiv preprint arXiv:1412.6572}, 2014.

\bibitem[\protect\citeauthoryear{Graham}{1972}]{Graham1972}
R.L. Graham.
\newblock An efficient algorithm for determining the con-vex hull of a finite
  planar set.
\newblock {\em Information Processing Letters}, 1(4):132 -- 133, 1972.

\bibitem[\protect\citeauthoryear{Huang \bgroup \em et al.\egroup
  }{2018}]{HuangKCH2018}
Chengqiang Huang, Yulei Wu, Geyong Min, and Yiming Ying.
\newblock Kernelized convex hull approximation and its applications in data
  description tasks.
\newblock In {\em 2018 International Joint Conference on Neural Networks
  (IJCNN)}, pages 1--8. IEEE, 2018.

\bibitem[\protect\citeauthoryear{Jarvis}{1973}]{Jarvis1973}
R.A. Jarvis.
\newblock On the identification of the convex hull of a finite set of points in
  the plane.
\newblock {\em Information Processing Letters}, 2(1):18--21, 1973.

\bibitem[\protect\citeauthoryear{Jia \bgroup \em et al.\egroup
  }{2019}]{CommunityGAN}
Yuting Jia, Qinqin Zhang, Weinan Zhang, and Xinbing Wang.
\newblock Communitygan: Community detection with generative adversarial nets.
\newblock {\em CoRR}, abs/1901.06631, 2019.

\bibitem[\protect\citeauthoryear{Krizhevsky}{2009}]{cifar-10}
Alex Krizhevsky.
\newblock Learning multiple layers of features from tiny images.
\newblock Technical report, Citeseer, 2009.

\bibitem[\protect\citeauthoryear{{Lecun} \bgroup \em et al.\egroup
  }{1998}]{mnist}
Y.~{Lecun}, L.~{Bottou}, Y.~{Bengio}, and P.~{Haffner}.
\newblock Gradient-based learning applied to document recognition.
\newblock {\em Proceedings of the IEEE}, 86(11):2278--2324, Nov 1998.

\bibitem[\protect\citeauthoryear{Lei \bgroup \em et al.\egroup
  }{2017}]{Jordan2017}
Lihua Lei, Cheng Ju, Jianbo Chen, and Michael~I Jordan.
\newblock Non-convex finite-sum optimization via scsg methods.
\newblock In {\em Advances in Neural Information Processing Systems}, pages
  2348--2358, 2017.

\bibitem[\protect\citeauthoryear{Li \bgroup \em et al.\egroup
  }{2019}]{Jianli2019Archive}
Jian Li, Xuanyuan Luo, and Mingda Qiao.
\newblock On generalization error bounds of noisy gradient methods for
  non-convex learning.
\newblock {\em arXiv preprint arXiv:1902.00621}, 2019.

\bibitem[\protect\citeauthoryear{Lov{\'a}sz and Vempala}{2006}]{lovasz2003focs}
L{\'a}szl{\'o} Lov{\'a}sz and Santosh Vempala.
\newblock Simulated annealing in convex bodies and an o*(n4) volume algorithm.
\newblock {\em Journal of Computer and System Sciences}, 72(2):392--417, 2006.

\bibitem[\protect\citeauthoryear{Nelles}{2013}]{nelles2013nonlinear}
Oliver Nelles.
\newblock {\em Nonlinear system identification: from classical approaches to
  neural networks and fuzzy models}.
\newblock Springer Science \& Business Media, 2013.

\bibitem[\protect\citeauthoryear{Perozzi \bgroup \em et al.\egroup
  }{2014}]{deepwalk}
Bryan Perozzi, Rami Al-Rfou, and Steven Skiena.
\newblock Deepwalk: Online learning of social representations.
\newblock In {\em Proceedings of the 20th ACM SIGKDD international conference
  on Knowledge discovery and data mining}, pages 701--710. ACM, 2014.

\bibitem[\protect\citeauthoryear{Raghu \bgroup \em et al.\egroup
  }{2017a}]{raghu2017svcca}
Maithra Raghu, Justin Gilmer, Jason Yosinski, and Jascha Sohl-Dickstein.
\newblock Svcca: Singular vector canonical correlation analysis for deep
  learning dynamics and interpretability.
\newblock In {\em Advances in Neural Information Processing Systems}, pages
  6076--6085, 2017.

\bibitem[\protect\citeauthoryear{Raghu \bgroup \em et al.\egroup
  }{2017b}]{Raghu2017Expressive}
Maithra Raghu, Ben Poole, Jon Kleinberg, Surya Ganguli, and Jascha~Sohl
  Dickstein.
\newblock On the expressive power of deep neural networks.
\newblock In {\em Proceedings of the 34th International Conference on Machine
  Learning-Volume 70}, pages 2847--2854. JMLR. org, 2017.

\bibitem[\protect\citeauthoryear{Sartipizadeh and
  Vincent}{2016}]{HosseinACH2016}
Hossein Sartipizadeh and Tyrone~L Vincent.
\newblock Computing the approximate convex hull in high dimensions.
\newblock {\em arXiv preprint arXiv:1603.04422}, 2016.

\bibitem[\protect\citeauthoryear{Schmidhuber}{2015}]{schmidhuber2015deep}
J{\"u}rgen Schmidhuber.
\newblock Deep learning in neural networks: An overview.
\newblock {\em Neural networks}, 61:85--117, 2015.

\bibitem[\protect\citeauthoryear{van~der Maaten}{2014}]{Maaten2014}
Laurens van~der Maaten.
\newblock Accelerating t-sne using tree-based algorithms.
\newblock {\em Journal of Machine Learning Research}, 15:3221--3245, 2014.

\bibitem[\protect\citeauthoryear{Wang \bgroup \em et al.\egroup
  }{2018a}]{GraphGAN}
Hongwei Wang, Jia Wang, Jialin Wang, Miao Zhao, Weinan Zhang, Fuzheng Zhang,
  Xing Xie, and Minyi Guo.
\newblock Graphgan: Graph representation learning with generative adversarial
  nets.
\newblock In {\em AAAI Conference on Artificial Intelligence}, 2018.

\bibitem[\protect\citeauthoryear{Wang \bgroup \em et al.\egroup
  }{2018b}]{wang2018shine}
Hongwei Wang, Fuzheng Zhang, Min Hou, Xing Xie, Minyi Guo, and Qi~Liu.
\newblock Shine: signed heterogeneous information network embedding for
  sentiment link prediction.
\newblock In {\em Proceedings of the Eleventh ACM International Conference on
  Web Search and Data Mining}, pages 592--600. ACM, 2018.

\bibitem[\protect\citeauthoryear{Wang \bgroup \em et al.\egroup
  }{2018c}]{wang2018towards}
Liwei Wang, Lunjia Hu, Jiayuan Gu, Zhiqiang Hu, Yue Wu, Kun He, and John
  Hopcroft.
\newblock Towards understanding learning representations: To what extent do
  different neural networks learn the same representation.
\newblock In {\em Advances in Neural Information Processing Systems}, pages
  9607--9616, 2018.

\bibitem[\protect\citeauthoryear{Wang \bgroup \em et al.\egroup
  }{2018d}]{acekg}
Ruijie Wang, Yuchen Yan, Jialu Wang, Yuting Jia, Ye~Zhang, Weinan Zhang, and
  Xinbing Wang.
\newblock Acekg: A large-scale knowledge graph for academic data mining.
\newblock In {\em Proceedings of the 27th ACM International Conference on
  Information and Knowledge Management}, CIKM '18, pages 1487--1490, New York,
  NY, USA, 2018. ACM.

\bibitem[\protect\citeauthoryear{Yu \bgroup \em et al.\egroup
  }{2018}]{yu2018curvature}
Tao Yu, Huan Long, and John~E Hopcroft.
\newblock Curvature-based comparison of two neural networks.
\newblock In {\em 2018 24th International Conference on Pattern Recognition
  (ICPR)}, pages 441--447. IEEE, 2018.

\end{thebibliography}

\end{document}